\documentclass[10pt,twocolumn,letterpaper]{article}

\usepackage{cvpr}              
\usepackage{times}
\usepackage{epsfig}
\usepackage{graphicx}
\usepackage{amsmath}
\usepackage{amssymb}
\usepackage{booktabs}
\usepackage[colorlinks,linkcolor=red,anchorcolor=green,citecolor=blue]{hyperref}
\usepackage[capitalize]{cleveref}
\usepackage[linesnumbered,ruled,vlined]{algorithm2e}
\usepackage{url}
\usepackage[utf8]{inputenc} 
\usepackage[T1]{fontenc}    
\usepackage{amsfonts}       
\usepackage{nicefrac}       
\usepackage{microtype}      
\usepackage{wrapfig}
\usepackage{multirow}
\usepackage{dsfont}
\usepackage{setspace}
\usepackage{caption}
\usepackage{amssymb}
\usepackage{fontawesome}
\usepackage{bbm}
\newcommand*{\method}{Detective}
\usepackage{colortbl}
\definecolor{myyellow}{rgb}{1,1, 0.6}
\definecolor{myorange}{rgb}{1, 0.8, 0.6}
\definecolor{myred}{rgb}{1, 0.6, 0.6}
\crefname{section}{Sec.}{Secs.}
\Crefname{section}{Section}{Sections}
\Crefname{table}{Table}{Tables}
\crefname{table}{Tab.}{Tabs.}
\usepackage{enumitem}

\usepackage[dvipsnames]{xcolor}

\definecolor{cvprblue}{rgb}{0.21,0.49,0.74}

\def\paperID{2222} 
\def\confName{CVPR}
\def\confYear{2024}

\title{Revisiting the Domain Shift and Sample Uncertainty in \\ Multi-source Active Domain Transfer}
\author{
Wenqiao Zhang\\
Zhejiang University\\
{\tt\small wenqiaozhang@zju.edu.cn}
\and
Zheqi Lv\\
Zhejiang University\\
{\tt\small zheqilv@zju.edu.cn}
\and
Hao Zhou\\
Harbin Institute of Technology\\
{\tt\small 2021210665@stu.hit.edu.cn}
\and
Jia-Wei Liu\\
National University of Singapore\\
{\tt\small jiawei.liu@u.nus.edu}
\and
Juncheng Li\\
Zhejiang University\\
{\tt\small junchengli@zju.edu.cn}
\and
Mengze Li\\
Zhejiang University\\
{\tt\small mengzeli@zju.edu.cn}
\and
\and
Siliang Tang\\
Zhejiang University\\
{\tt\small siliang@zju.edu.cn}
\and
Yueting Zhuang\\
Zhejiang University\\
{\tt\small yzhuang@zju.edu.cn}
}

\begin{document}
\maketitle
\begin{abstract}
\label{sec:abstract}
Active Domain Adaptation (ADA) aims to maximally boost model adaptation in a new target domain by actively selecting a limited number of target data to annotate. This setting neglects the more practical scenario where training data are collected from multiple sources. This motivates us to target a new and challenging setting of knowledge transfer that extends ADA from a single source domain to multiple source domains, termed Multi-source Active Domain Adaptation (MADA). Not surprisingly, we find that most traditional ADA methods cannot work directly in such a setting, mainly due to the excessive domain gap introduced by all the source domains and thus their uncertainty-aware sample selection can easily become miscalibrated under the multi-domain shifts.
Considering this, we propose a \underline{\textbf{D}}ynamic int\underline{\textbf{e}}gra\underline{\textbf{te}}d un\underline{\textbf{c}}er\underline{\textbf{t}}a\underline{\textbf{i}}nty \underline{\textbf{v}}aluation fram\underline{\textbf{e}}work~(\textbf{Detective}) that comprehensively consider the domain shift between multi-source domains and target domain to detect the informative target samples. Specifically, the \method{} leverages a dynamic Domain Adaptation (DA) model that learns how to adapt the model’s parameters to fit the union of multi-source domains. This enables an approximate single-source domain modeling by the dynamic model. We then comprehensively measure both domain uncertainty and predictive uncertainty in the target domain to detect informative target samples using evidential deep learning, thereby mitigating uncertainty miscalibration. Furthermore, we introduce a contextual diversity-aware calculator to enhance the diversity of the selected samples. Experiments demonstrate that our solution outperforms existing methods by a considerable margin on three domain adaptation benchmarks.



\end{abstract}

\section{Introduction}
\label{sec:introduction}

Unsupervised Domain Adaptation (UDA) aims to transfer knowledge learned from labeled data in the original domain (source $\mathcal{D}_s$) to the new domain (target $\mathcal{D}_t$). Although UDA is capable of alleviating the poor generalization of learned deep neural networks when the data distribution significantly deviates from the original domain~\cite{wang2018deep,you2019universal,tzeng2017adversarial}, the unavailability of target labels greatly hinders its performance. This presents a significant gap compared to its supervised counterpart~\cite{tsai2018learning,chen2018domain}. An appealing way to address this issue is by actively collecting informative target samples within an acceptable budget, thereby maximally benefiting the adaptation model. This promising adaptation paradigm integrates the idea of active learning~\cite{settles2009active} into traditional UDA, known as Active Domain Adaptation (ADA)~\cite{rai2010domain}.

\begin{figure*}[t]
\vspace{-0.3cm}
\includegraphics[width=0.91\textwidth]{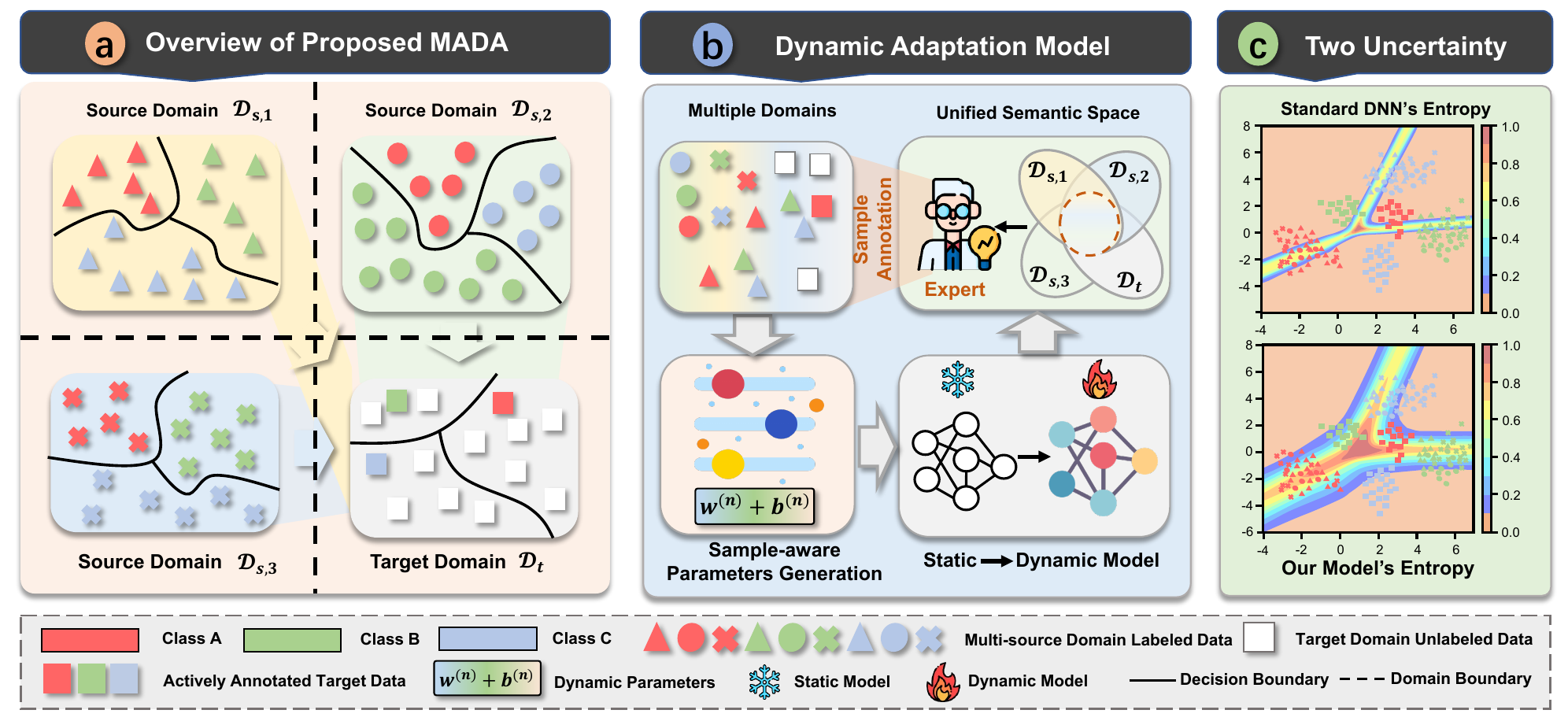}
 \vspace{-0.2cm}
\centering\caption{
(a) illustrates an overview of the proposed MADA.
(b) depicts a concise version of the dynamic adaptation and sample selection strategy.
(c) compares the entropy of general DNNs with the entropy of our model, where both models are trained using source data.
}
\vspace{-0.3cm}
\label{intro}
\end{figure*}

While the existing ADA framework~\cite{rai2010domain,su2020active,fu2021transferable,prabhu2021active,de2021discrepancy} presumes that all labeled training data share the same distribution from a single source domain $\mathcal{D}_s$, it conceals the practicality that data is typically collected from various domains ($\{\mathcal{D}_{s,i}\}^M_{i=1}$, with $M$ representing the number of domains) in real-world scenarios~\cite{zhao2018adversarial}.
This setting hinders the model's ability to learn from diverse domains, one of the most precious capacity of human that adapting knowledge acquired from varied environments to unseen fields. With this consideration, we introduce a challenging and realistic problem setting termed Multi-source Active Domain Adaptation (MADA). 
As captured in Figure~\ref{intro}(a),  MADA posits the availability of several labeled source domains, capable of annotating a small portion of valuable target samples for maximally benefiting the adaptation process. 
The proposed MADA, while straightforward and promising, encounters multiple challenges when directly applying conventional ADA techniques to MADA task:
\textbf{i}) \textbf{Multi-grained Domain Shift.} In ADA, acquiring target labels involves gauging the domain shift between a single $\mathcal{D}_s$ and the target domain $\mathcal{D}_t$ to identify informative target samples.
However, the significantly variant data distributions of $\{\mathcal{D}_{s,i}\}^M_{i=1}$ also pose challenges for MADA, 
as it encompasses multiple source domains with disparate distributions, \emph{i.e.}, the alignment among multiple source domains should be carefully addressed for a reliable MADA model. 
\textbf{ii}) \textbf{Uncertainty Miscalibration.} The uncertainty-aware selection criteria in current ADA methods typically rely on predictions from deterministic models, which are prone to miscalibration when faced with data exhibiting distribution shifts~\cite{guo2017calibration}. Relying solely on predictive uncertainty proves to be unreliable, as standard DNNs often give misplaced overconfidence in their predictions on target data (Figure~\ref{intro}(c)), potentially impairing MADA's performance. In summary, these shortcomings necessitate a thorough reevaluation of both MADA and its corresponding solutions.

To alleviate the aforementioned limitations, we propose a 
\underline{\textbf{D}}ynamic int\underline{\textbf{e}}gra\underline{\textbf{te}}d un\underline{\textbf{c}}er\underline{\textbf{t}}a\underline{\textbf{i}}nty \underline{\textbf{v}}aluation fram\underline{\textbf{e}}work, abbreviated as \textbf{\method{}}. Detective comprises the following components: The first module, termed the Universal Dynamic Network (UDN), utilizes a Hypernetwork~\cite{ha2016hypernetworks} to derive a domain-agnostic model spanning $\{\mathcal{D}_{s,i}\}^M_{i=1}$. As illustrated in Figure~\ref{intro}(b), the key insight is to fit the model towards the union space of $\{\mathcal{D}_{s,i}\}^M_{i=1}$ by adapting its parameters.
Consequently, the model is divided into static layers (backbone) and dynamic layers (classifier). The static layers maintain fixed parameters, whereas the dynamic layers' parameters (classifier) are dynamically generated depending on multiple source samples. Employing this adaptable domain-agnostic model allows us to treat $\{\mathcal{D}_{s,i}\}^M_{i=1}$ as one-source domain. This significantly simplifies the alignment process between $\{\mathcal{D}_{s,i}\}^M_{i=1}$ and $\mathcal{D}_t$, as it is no longer necessary to pull all source samples together with the target samples.
Next, we develop the integrated uncertainty selector (IUS) derived from Evidential Deep Learning (EDL)~\cite{sensoy2018evidential} to effectively measure the domain's characteristic to the target domain. In EDL, categorical predictions are construed as distributions, with a Dirichlet prior~\cite{sethuraman1994constructive} applied to class probabilities, transforming the prediction from a point estimator to a probabilistic one.
We regard the EDL's prediction as the \emph{domain uncertainty} as it enables the detection of unfamiliar data instances~\cite{nandy2020towards} illustrated in Figure~\ref{intro}(c), \emph{i.e.}, the miscalibration of deterministic's prediction can be mitigated by considering the spectrum of possible outcomes.
By amalgamating the predictive uncertainty from a standard DNN, we integrate both domain and predictive uncertainties to select samples that are informative for the target domain. Moreover, we  design a contextual diversity calculator (CDC) that evaluates the diversity of chosen target samples at the image-level, guaranteeing the information diversity of selected samples.


To summarize, our contributions are fourfold:
(1) We have designed a practical yet challenging task, Multi-source Active Domain Adaptation (MADA). This task aims to transfer knowledge from multiple labeled source domains to a target domain with a small amount of actively labeled samples.
(2) We propose a universal dynamic network capable of adaptively mapping samples from multi-source domains to the adaptation model's parameters. This effectively alleviates the domain shift between the multi-source domains and the target domain.
(3) We introduce a novel integrated uncertainty calculation strategy to tackle the MADA challenge. This strategy thoroughly assesses both domain and predictive uncertainties to select informative samples and is complemented by a contextual diversity calculator to enrich the information diversity of the selected data.
(4) Experimental studies demonstrate that our method significantly improves upon the prevalent ADA methodologies for MADA.

\section{Related Work}
\label{sec:related_work}
\noindent\textbf{Active\&Multi Domain Adaptation.} 
To improve the generalization of the model, domain adaptation is receiving more and more attention from researchers \cite{li2023unsupervised, prabhu2021active}.
Active Domain Adaptation (ADA) adapts a source model to an unlabeled target domain by using an oracle to obtain the labels of selected target instances. ALDA~\cite{rai2010domain} samples instances based on model uncertainty and learned domain separators and applies them to text data sentiment classification. ~\cite{su2020active} and ~\cite{fu2021transferable} introduce ADA into adversarial learning and identify domains through the resulting domain discriminators. However, they may give the same high score for most of the target data, so it is not sufficient to ensure that the selected samples represent the entire target distribution. ~\cite{prabhu2021active,de2021discrepancy} improves this weakness by selecting active samples by clustering, but they still focus on the measurement of prediction uncertainty like existing ADA methods, which leads to sometimes being misunderstood on the target data.
Multi-source domain adaptation(MSDA) aims to learn domain-invariant features across all domains, or leverage auxiliary classifiers trained with multi-source domain to ensemble a robust classifier for the target domain~\cite{sun2015survey,sun2011two}.
~\cite{li2020online} presents a framework to search for the best initial conditions for MSDA via meta-learning. ~\cite{li2021dynamic} propose dynamic transfer which adapts the model’s parameters for each sample to simplify the  alignment between source domains and target domain and address domain conflicts.
MDDA~\cite{zhao2020multi} considers the distances on not only the domain level but also the sample level, it selects source samples which is similar to the target to finetune the source-specific classifiers.
In addition, multi-type data domain adaptation and fusion have also been greatly developed \cite{zhang2019frame, li2022end, zhang2021consensus, li2022hero, zhang2022magic}.
Despite promising, the aforementioned methods can not directly transfer to the MADA, due to they fail to measure the multi-source domain to select valuable target samples with multi-grained domain shift. 

\noindent\textbf{HyperNetwork.} 
HyperNetwork~\cite{ref:hypernetworks} originally aimed to achieve model compression by reducing the number of parameters a model needs to train. It then develops into a neural network that generates parameters for another neural network. Due to its fast and powerful advantages in generalization, research on HyperNetwork has gradually increased.
In terms of the theory, ~\cite{ref:hypernetwork_initial} studies the parameter initialization of HyperNetwork. In terms of application, HyperNetwork research is widely used in various tasks, such as few-shot learning~\cite{ref:hypernetwork_one_shot}, continual learning~\cite{ref:hypernetwork_continual_learning}, graph learning~\cite{ref:hypernetwork_graph}, recommendation system~\cite{ref:hypernetwork_apg}, device-cloud collaboration~\cite{ref:hypernetwork_duet,hypernetwork_ideal}, etc.
To the best of our knowledge, we are the first to leverage HyperNetwork to the adaptation problem with unique challenges arising from the problem setup.

\noindent\textbf{Evidential Deep Learning (EDL).} 
Although deep learning models have made surprising progress \cite{zhang2022boostmis, li2023winner, zhang2023learning, li2023multi,  li2020ib}, how to effectively evaluate the uncertainty of model predictions is still worth pondering.
~\cite{sensoy2018evidential} first proposed to estimate reliable classification uncertainty via EDL. It places Dirichlet priors on class probabilities to interpret classification predictions as distributions and achieves significant advantages in out-of-distribution queries.
As the potential of EDL is gradually being explored, 
~\cite{amini2020deep} introduces the evidential theory to regression tasks by placing evidential priors during training such that the model is regularized when its predicted evidence is not aligned with the correct output. 
~\cite{bao2021evidential} propose a novel model calibration method to regularize the EDL training to estimate uncertainty for open-set action recognition. ~\cite{chen2022evidential} introduces a class competition-free uncertainty score based on EDL to find potential unknown samples in universal domain adaptation. 
Differently, we introduce the EDL theory into MADA that effectively measures the domain uncertainty to boost the adaptation.

\section{Methodology}
\label{sec:methodology}

\subsection{Problem Formulation}
Formally, we have access to $M$ fully labeled source domains and a target domain with actively labeled target data within a pre-defined budget $\mathcal{B}$ in MSDA. The $i$-th source domain $\mathcal{D}_{s,i}=\{(x^{(j)}_{s,i}, y^{(j)}_{s,i})\}_{j=1}^{N_{s,i}}$ contains $N_{s,i}$ labeled data sampled from the source distribution $P_{s,i}(x, y)$, where $i \in M$. The target domain $\mathcal{D}_{t}$  contains unlabeled samples $\{x^{(j)}_{tu}\}_{j=1}^{N_{tu}}$ from the target distribution $P_{t}(x, y)$. Following the standard ADA setting~\cite{xie2022active}, the size of budget $\mathcal{B}$ set to $N_{tl}$, the labeled target domain $\mathcal{D}_{tl}$=$\{x^{(j)}_{tl}\}_{j=1}^{N_{tl}}$, where $N_{tl} \ll N_{tu}$ and $N_{tl} \ll N_{s,i}$.
$P_{s,i}(x, y) \neq P_{t}(x, y)$ and $P_{s,i}(x, y) \neq P_{s,j}(x, y)$ where $i \neq j$. The multiple source domains and target domain have the same label space $Y$ = $\{1, 2, \cdots, K\}$ with $K$ categories. We aim to learn an model $\mathcal{M}(\cdot)$ adapting from $\{\mathcal{D}_{s,i}\}^M_{i=1}$ to $\mathcal{D}_t$, \emph{i.e.}, the model can generalize well on unseen samples from $\mathcal{D}_t$. In general, $\mathcal{M}(\cdot)$ consists of two functions as below:
\begin{equation}
\begin{aligned}
&\underbrace{\mathcal{M}(\cdot;(\Theta_{{m}},\Theta_{{a}}))}_{\rm{MADA\ Model}}: \underbrace{\mathcal{M}_m((\{\mathcal{D}_{s,i}\}^M_{i=1},\mathcal{D}_{tl});\Theta_{m})}_{\rm{Multi-domain\ Adaptation\ Model}} \\& \leftrightarrow
\underbrace{\mathcal{M}_a((\mathcal{D}_{tl}|\mathcal{D}_{t});\Theta_{{a}})}_{\rm{Active\ Learning\ Model}}\,,
 \label{adv}
\end{aligned}
\end{equation}
where $\mathcal{M}_m(\cdot;\Theta_{m})$ is the multi-domain learning model and $\mathcal{M}_a(\cdot;\Theta_{a})$ is the annotation candidate selection through active learning model with parameters $\Theta_{m}$ and $\Theta_{a}$.

\begin{figure*}[t]
\vspace{-0.3cm}
\includegraphics[width=0.95\textwidth]{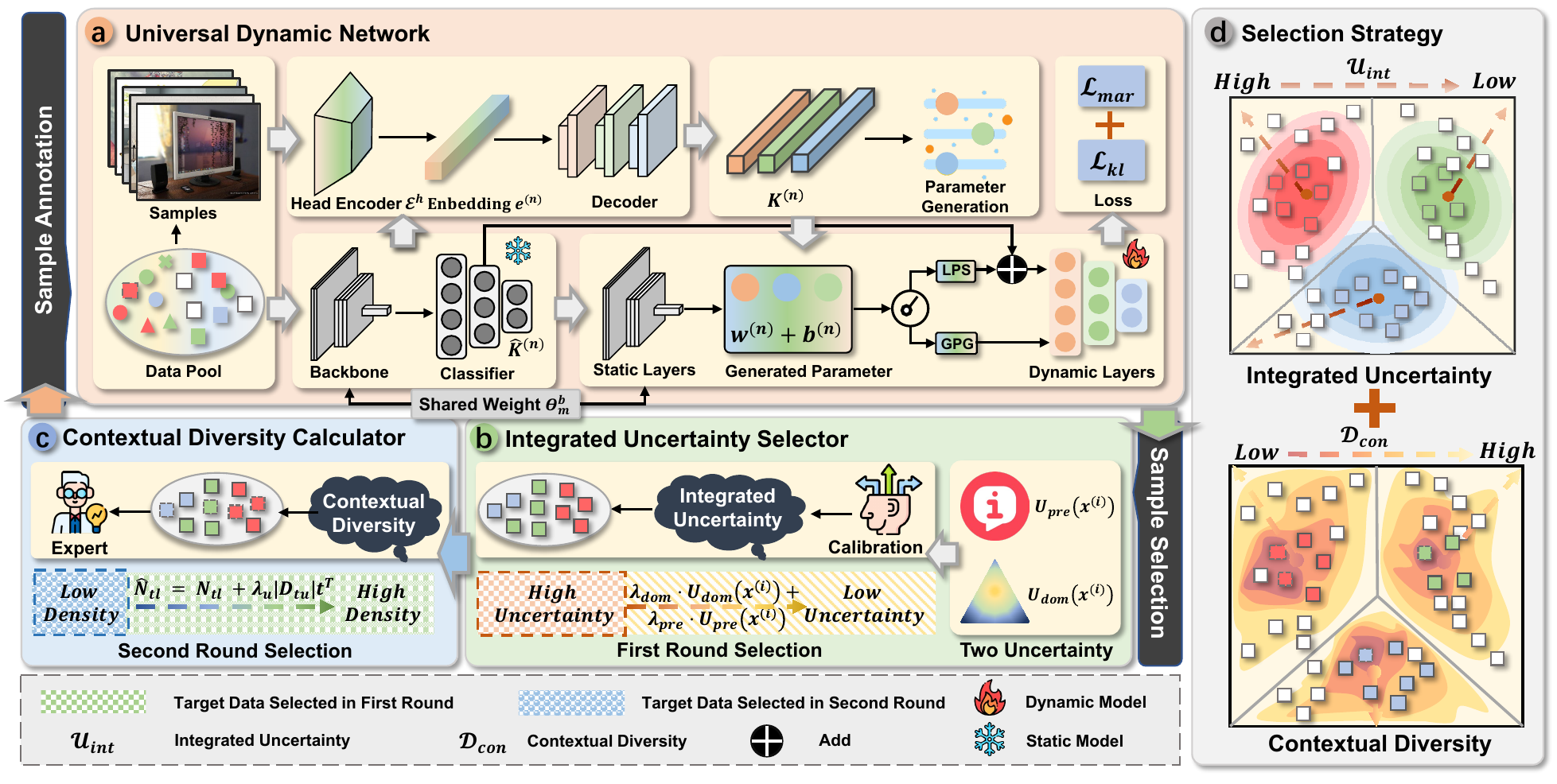}
\vspace{-0.3cm}
\centering\caption{\textbf{Overview of \method{}}. \textbf{(a)} Universal Dynamic Network (UDN) leveraging the Hypernetworks~\cite{ha2016hypernetworks} that can generate the dynamic parameters for the adaptive classifiers conditioned on the different domain samples. \textbf{(b)} and \textbf{(c)} are the Integrated Uncertainty Selector (IUS) and Contextual Diversity Calculator (CDC) that respectively measure the uncertainty (first round selection) and diversity (second round selection) of target samples. \textbf{(d)} visualizes the two active learning strategies. }
\vspace{-0.2cm}
\label{framework}
\end{figure*}
\subsection{Universal Dynamic Network}
The Universal Dynamic Network (UDN) (Figure~\ref{framework}(a)) can generate the dynamic parameters for the adaptive classifier conditioned on input samples from different domains, which elegantly regard the multi-domains as one single domain. We train a primary model with a backbone and a classifier for developing the global adaptation model. Given the $M$ source domain samples $\{\{(x^{(j)}_{s,i}, y^{(j)}_{s,i})\}_{j=1}^{N_{s,i}}\}_i^{M}$, the proposed UDN can be described as below:
 \begin{equation}
 \begin{aligned}
 P (p|x(i))= \Psi (\Omega(x^{(j)}_{s,i};\Theta_m^b); \Theta_m^c)\,,
    \label{eq:umn}
\end{aligned}
\end{equation}
where $\Omega(·;\Theta_m^b)$ is the backbone extracting features from input samples and $\Psi(·;\Theta_m^c)$ is the classifier. $\Theta_m^b$ and $\Theta_m^c$ are the learnable parameters for the backbone and classifier.

Here, we treat the backbone as ``static layers'' and the classifier as ``dynamic layers'' to achieve dynamic adaptation: 
\begin{itemize}
[topsep=0pt,leftmargin=20pt]
\item \textbf{Static Layers.} The backbone with $\Theta_m^b$ learned from multi-domain data can accurately map the image into the feature space. We fixed the backbone as ``static layers'' to generate a generalized representation for any given sample concerning the multi-domain distribution. 
\item \textbf{Dynamic Layers.} Depending on the image features, the dynamic layers learn adaptive classifier weights ${\Theta}_m^c$. It learns to adapt the parameters to fit
the model to the distribution formed by the union of source domains.
\end{itemize}



\label{sec:udn}
\noindent\textbf{Retrospect of HyperNetwork.} 
\label{sec:hyp}
The core of dynamic layers are based on the HyperNetwork~\cite{ha2016hypernetworks}, we will outline the procedure for using a HyperNetwork to generate the parameters for another neural network.
Specifically, HyperNetwork treats the parameters of the multi-layer perception (MLP) as a matrix $K^{(n)} \in \mathbb{R}^{N_{in}\times N_{out}}$, where $N_{in}$ and $N_{out}$ represent the number of input and output neurons of the $n^{th}$ layer of MLP, respectively. $N_{in}$ and $N_{out}$ portray the structure of the MLP layers together. The generation process of $K^{(n)}$ can be regarded as a matrix factorization as below:
 \begin{align}
    K^{(n)} = \xi(z^{(n)};\Theta_p), \forall n=  1, \cdots, N_l\,,
\label{eq:hyper}
\end{align}
where $z^{(n)}$ and $\xi(\cdot)$ are randomly initialized with parameters $\Theta_p$ in training procedure. 
The gradients are backpropagated to $z^{(n)}$ and $\xi(\cdot)$, which can help to update them. $z^{(n)}$ and $\xi(\cdot)$ will be saved instead of $K^{(n)}$.

\begin{sloppypar} 
\noindent\textbf{HyperNetwork-based Dynamic Layers.} However, as shown in Eq.(\ref{eq:hyper}), the original HyperNetwork only utilizes a randomly initialized $z^{(n)}$ to generate the parameters, which lacks the interaction between the parameter generation process and the current input. To measure the dynamic layers for different domains, we propose to model the parameters by replacing the $z^{(n)}$  with representations of the current input image. Specifically, given the feature $\Omega(x^{(i)};\Theta_m^b)$ extracted from backbone of sample $x^{(j)}_{s,i}$, we first develop a layer-specific encoder $\mathcal{E}^{h}(\cdot)$ that encodes the $\Omega(x^{(i)};\Theta_m^b)$ as $\boldsymbol{e}^{h}$. Then the HyperNetwork is used to convert the embedding $\textbf{e}^{(n)}$ into parameters, \emph{i.e.}, we input $\textbf{e}^{(n)}$ into the following two MLP layers to generate parameters of dynamic layers:
\end{sloppypar}
\vspace{-0.3cm}
\begin{align}
\label{eq:kernal_generation_detail}
\begin{gathered}
\mathcal{\boldsymbol{w}}^{(n)} = (W_1\mathcal{E}^{h}(\Omega(x^{(i)};\Theta_m^b)) + B_1)W_2 + B_2,
\end{gathered}
\end{align}
where the first and second MLP layers are respectively defined by their weights, $W_1$ and $W_2$, along with bias terms $B_1$ and $B_2$, encapsulating their unique characteristics.



By doing so, we present two dynamic layer learning strategies, local parameters shift (\textbf{LPS}) and global parameters generation (\textbf{GPG}). The LPS adopts a two-stage training strategy, \emph{i.e.}, we first pretrain the classifier $\Psi(·;\Theta_m^c)$ together with static layers, $\hat{K}^{(n)}$ denote the parameters for n-th layer. The matrices $\hat{K}^{(n)}$ can be seen as a basis for classifier weight space, although they are not necessarily linearly independent. Then we finetune the classifier with local parameters shift, which can be seen as the projections of the residual matrix in the corresponding weight subspaces. For GPG, the parameters are directly generated by the sample-aware learning manner, together with the static layers that are conditioned on the multi-source domain data.

\begin{equation} \!\!\!\!K^{(n)} \!\! = \!\!
\begin{cases} 
\hat{K}^{(n)} +\mathcal{\boldsymbol{w}}^{(n)}  + \mathcal{\boldsymbol{b}}^{(n)},& \!\!\!\!\!\mbox{if } \Psi(·;\Theta_m^c) \mbox{  pretrained}\\
\mathcal{\boldsymbol{w}}^{(n)}  + \mathcal{\boldsymbol{b}}^{(n)}, & \!\!\!\!\! \mbox{otherwise }
\end{cases} 
\label{computeA}
\end{equation}
where $\forall n=  1, \cdots, N_l$, $K^{(n)}$ denotes the $n^{th}$ layer parameters of dynamic layers.

In summary, the UDN $\mathcal{M}_m(\cdot;\Theta_m)$ learns to adapt the parameters to fit the model to the distribution formed by the union of source domains. The target domain is not required to be aligned with any specific domains and there are no rigid domain boundaries. The model parameters $\Theta_m$ can be similar for examples from different domains and different for examples from the same domain.

\subsection{Integrated Uncertainty Selector}
\noindent\textbf{Evidential Deep Learning with DNN.} 
 The above learned multi-domain adaptation model $\mathcal{M}_m(\cdot;\Theta_m)$ explicitly modifies the model's weights according to the samples' distribution from different domains. However, this model has a vague understanding of the target data distribution for out-of-source domain. Therefore, we develop the Integrated Uncertainty Selector to choose the informative samples to facilitate target domain adaptation. Traditional active learning often adopts the prediction of class probability $p$ vector to measure the uncertainty as the criteria of sample selection. For active domain adaptation (ADA), such a manner essentially gives a point estimate of $p$ and can be easily miscalibrated on data with distribution shift~\cite{guo2017calibration}. Motivated by Evidential Deep Learning (EDL)~\cite{sensoy2018evidential}, unlike the traditional active learning, we predict a high-order Dirichlet distribution $Dir(p|\alpha)$ which is a conjugate prior of the lower-order categorical likelihood. Specifically, a Dirichlet distribution 
 is placed over $p$ to represent the probability density of each variable $p$. Given sample $x^{(i)}$, the probability density function of $p$ is denoted as $P(p|x^{(i)})=Dir(p|\alpha^{(i)})$.
\begin{equation}
\begin{aligned}
Dir(p|\alpha^{(i)})=
 \begin{cases} 
\frac{\Gamma (\sum_{d=1}^D \alpha_{d}^{(i)})}{\prod_{d=1}^D\Gamma(\alpha_{d}^{(i)})}\prod_{d=1}^D p_{d}^{\alpha_{d}^{(i)}-1}, \!\! & \mbox{if } p \in \Delta^D\\
\quad\quad\quad\quad\quad 0 \quad\quad\quad\quad\quad, & \mbox{otherwise}
\end{cases} 
 \label{edl}
\end{aligned}
\end{equation}
where $\alpha_{d}^{(i)}$ is the parameters of the Dirichlet distribution for sample $x^{(i)}$, $\Gamma(\cdot)$ is the Gamma function and $\Delta^D$ the $D$-dimensional unit simplex, where $\Delta^D$=$\{\sum_{d=1}^D p_{d}=1 \& 0\leq p_d \leq 1 $\}, $\alpha^{(i)}=g(f(x^{(i)},\Theta))$, and $g(\cdot)$ is a function (\emph{e.g.}, exponential function) to keep $\alpha^{(i)}$ positive. In this way, the prediction of each sample is interpreted as a distribution over the probability simplex, rather than the simple predictive uncertainty. And we can mitigate the uncertainty miscalibration by considering all possible predictions rather than unilateral predictions.

After applying the EDL to a standard DNN with softmax for the classification task, the predicted probability for class $k$ can be denoted as Eq.~(\ref{edl}), by marginalizing over $p$. The details of derivation in the supplementary material.
\begin{equation}
\begin{aligned}
&\hat{P}(y=k|x^{(i)})= \int P(y=k|p)P(p|x^{(i)})dp\\&=\frac{g(f_k(x^{(i)}))}{\sum_{c=1}^K g(f_c(x^{(i)}))}=\mathbbm{E}[Dir(p_k|\alpha^{(i)})]\,,
 \label{edl}
\end{aligned}
\end{equation}
where $\hat{P}$ is the prediction. If $g(\cdot)$ adopts the exponential function, then softmax-based DNNs can be regarded as predicting the expectation of Dirichlet distribution. 

\noindent\textbf{Domain \& Predictive Uncertainty Integration.} For the evidential model supervised with multi-source data, if target samples are obviously distinct from the source domain, \emph{e.g.}, the realistic \emph{v.s.} clipart style, the evidence collected for these target samples may be insufficient, because the model lacks the knowledge about this kind of data. To solve this problem, we use the uncertainty obtained from the lack of evidence, \emph{i.e.}, domain uncertainty, to measure the target domain's characteristics. Specifically, the domain uncertainty (Figure~\ref{framework}(d)) $\mathcal{U}_{dom}$ of sample $x^{(i)}$ is defined as:
\begin{equation}
\begin{aligned}
&\mathcal{U}_{dom}(x^{(i)})= \sum_{k=1}^K \hat{P}(y=k|x^{(i)}) (\Phi(\alpha_{k}^{(i)}+1)- \\& \!\!\!\!\Phi(\sum_{c=1}^K \alpha_{c}^{(i)}+1))\!\!-\!\!\sum_{k=1}^K \hat{P}(y=k|x^{(i)}) {\rm log} \hat{P}(y=k|x^{(i)})\!\!\,,
 \label{domain}
\end{aligned}
\end{equation}
where $\Phi$ is the digamma function. Here, we use mutual information to measure the spread of Dirichlet distribution on
the simplex like~\cite{sensoy2018evidential}. Higher $\mathcal{U}_{dom}(x^{(i)})$ indicates larger domain uncertainty, \emph{i.e.}, the Dirichlet distribution is broadly spread on the probability simplex.

We also utilize the predictive entropy to quantify predictive uncertainty, which is denoted as the expected entropy of all possible predictions. Specifically, given sample $x^{(i)}$, the predictive uncertainty (Figure~\ref{framework}(d)) $\mathcal{U}_{pre}(x^{(i)})$ is:
\begin{equation}
\begin{aligned}
&\mathcal{U}_{pre}(x^{(i)})= \mathbbm{E}[H[P(y|p)]]= \\& \sum_{k=1}^{K} P(y=k|x^{(i)}) (\Phi(\sum_{c=1}^{K}\alpha_{c}^{(i)}+1)-\Phi(\alpha_{k^{(i)}}+1))\,.
 \label{prediction}
\end{aligned}
\end{equation}

With the domain uncertainty $\mathcal{U}_{dom}$ and predictive uncertainty $\mathcal{U}_{pre}$, we comprehensively consider the two uncertainties to select the target domain samples. We use the mix-up strategy (Figure~\ref{framework}(b)) to fuse the $\mathcal{U}_{dom}$ and $\mathcal{U}_{pre}$:
$\mathcal{U}_{int}(x^{(i)})=\lambda_{dom}\cdot\mathcal{U}_{dom}(x^{(i)}) + \lambda_{pre} \cdot \mathcal{U}_{pre}(x^{(i)})\,,
 \label{prediction}$
where coefficients $\lambda_{dom}$ and $\lambda_{pre}$ are pre-defined hyperparameters. The bigger value of $\lambda_{dom}$ means the higher influence of domain uncertainty and vice versa.
\subsection{Contextual Diversity Calculator}
The selected target samples based on the aforementioned integrated uncertainty $\mathcal{U}_{int}$ may have similar semantics in the implicit feature space. We argue that when considering the potential similarity of selected samples in the learned feature space, not all selected samples contribute highly to the performance of MADA but instead increase the labeling cost and computational expense. To enhance the diversity of sample selection, we measure the contextual diversity (Figure~\ref{framework}(c))  based on feature space coverage. Specifically, we calculate the image-level density of selected samples $\mathcal{S}_t=\{x_{tu}^{(i)}\}_{j=1}^{J}$, the $j$-$th$  feature extracted by the backbone. Finally, we remove $\hat{{N}_{tl}}$ - ${{N}_{tl}}$ images with the largest image-level density value to reach the desired labeling budget ${{N}_{tl}}$. Note that ${{N}_{tl}}$ is the actual labeling budget and $\hat{\mathcal{N}_{tl}}$ is a hyperparameter. Obviously, the $\hat{{N}_{tl}}$ should be gradually increased during training, $\hat{{N}_{tl}}$ can be computed as:
\begin{equation}
\begin{aligned}
\hat{{N}_{tl}}={N}_{tl}+\lambda_u|\mathcal{D}_{tu}|t^{\tau}
 \label{prediction}
\end{aligned}
\end{equation}
where $\lambda_u$ is a hyperparameter. $t^{\tau}$ denotes its temperature. By modulating $t^{\tau}$, we can control the uncertainty-diversity tradeoff to further boost MADA performance.

\begin{table}
\begin{center}
\small
\captionsetup{font={small,stretch=1.25}, labelfont={bf}}
\caption{\textbf{Accuracy(\%) comparison on \texttt{Office-Home} using Resnet50 as backbone.}  Acronym of each model can be found in Section ~\ref{sec:setting}.
We color each row as the \colorbox{myred}{\textbf{best}}, \colorbox{myorange}{\textbf{second best}}, and \colorbox{myyellow}{\textbf{third best}}. $\cup$ represents the rest of the domains.  
}  
 \renewcommand{\arraystretch}{1.2}
 \resizebox{0.475\textwidth}{!}{
  \begin{tabular}{c|c||c c c c|c}
   \toprule[1.5pt]
    \multirow{2}{*}{\textbf{DA Setting}}&\multirow{2}{*}{\textbf{Methods}}  & \multicolumn{5}{c}{\texttt{Office-Home Dataset}} \\
    \cline{3-6} \cline{7-7}
    &  & \textbf{ $\cup \rightarrow$ A} & \textbf{$\cup  \rightarrow$  P} & \textbf{$\cup  \rightarrow$  C} & \textbf{$\cup  \rightarrow$ R}  & \textbf{Mean}\\\hline \hline
\multirow{4}{*}{MSDA} & MFSAN~\cite{zhu2019aligning}& 72.1& 80.3& 62.0& 81.8& 74.1\\
& MDDA~\cite{zhao2020multi}& 66.7& 79.5& 62.3& 79.6& 71.0\\
& SImpAI~\cite{venkat2020your}& 70.8& 80.2& 56.3& 81.5& 72.2\\
& MADAN~\cite{zhao2021madan}& 66.8& 78.2& 54.9& 81.5& 70.4\\
   \toprule[1pt]
\multirow{4}{*}{ADA} & CLUE~\cite{prabhu2021active}& 75.61 & 86.48& 67.84& 84.07 & 78.50 \\
   & TQS~\cite{fu2021transferable}& 78.29&	86.62	&68.74&	86.07&79.93
\\
   & DUC~\cite{xie2023dirichlet}&  \colorbox{myyellow}{\textbf{81.21}}& \colorbox{myorange}{\textbf{89.46}}& \colorbox{myyellow}{\textbf{71.63}}& \colorbox{myyellow}{\textbf{88.18}} & \colorbox{myyellow}{\textbf{82.62}}\\
   & EADA~\cite{xie2022active}& \colorbox{myorange}{\textbf{81.74}}& \colorbox{myyellow}{\textbf{89.15}}& \colorbox{myorange}{\textbf{71.75}}& \colorbox{myorange}{\textbf{89.18}}& \colorbox{myorange}{\textbf{83.03}} \\
   \toprule[1pt]
   \rowcolor{gray!40} MADA & \textbf{\method{} (Ours)} & \colorbox{myred}{\textbf{86.03}} & \colorbox{myred}{\textbf{91.78}} & \colorbox{myred}{\textbf{86.91}} & \colorbox{myred}{\textbf{95.22}} & \colorbox{myred}{\textbf{89.99}}\\ 
   \toprule[1.5pt]
  \end{tabular}
  }
  \label{tab:results_1}
\end{center}
\vspace{-0.3cm}
\end{table}

\subsection{Detective Training}
To get reliable and consistent opinions for labeled data, the evidential model is trained to generate
sharp Dirichlet distribution located at the corner of these labeled data. Concretely,
we train the model by minimizing the negative logarithm of the marginal likelihood ($\mathcal{L}_{mar}$) which is minimized to ensure the correctness of prediction.
\begin{equation}
\begin{aligned}
\mathcal{L}_{mar}=\!\!\!\!\!\sum_{x^{(i)} \in [\{\mathcal{D}_{s,i}\}^M_{i=1},\mathcal{D}_{t}]} \!\!\!\!\! \beta_{k}^{(i)} 
({\rm log} (\sum_{k=1}^k \alpha_k^{(i)})-log \alpha_k^{(i)})\,,
 \label{prediction}
\end{aligned}
\end{equation}
where [,] is the concatenation of two sets. $\beta_{k}^{(i)}$ is the $k$-th element of the one-hot label vector of sample $x^{(i)}$.

We also minimize the KL-divergence between two Dirichlet distributions ($\mathcal{L}_{kl}$) of source domains and target domain.
\begin{equation}
\begin{aligned}
\mathcal{L}_{kl}=\!\!\!\!\sum_{x^{(i)} \in [\{\mathcal{D}_{s,i}\}^M_{i=1},\mathcal{D}_{t}]} \!\!\!\!\! KL[Dir(p|\hat{\alpha^{(i)}})|Dir(p|1)]
 \label{prediction}
\end{aligned}
\end{equation}
where $\hat{\alpha^{(i)}}=\beta_{k}^{(i)}+(1-\beta_{k}^{(i)})\cdot\alpha^{(i)}$. $\hat{\alpha^{(i)}}$ can be seen as removing the evidence of ground-truth class. Minimizing $\mathcal{L}_{kl}$ will force the evidences of other classes to reduce, avoiding the collection of mis-leading evidences.

To sum up, the \method{} is learned with a combination of two losses:
$\mathcal{L}_{sum}= \lambda_{mar}\cdot\mathcal{L}_{mar}+\lambda_{kl}\cdot\mathcal{L}_{kl}\ $,
where $\lambda_{mar}$ and $\lambda_{kl}$ are hyperparameters. 

\section{Experiments}
\label{sec:experiments}

\begin{table}
    \begin{center}
     \caption{\textbf{Ablation study of the effect of individual module.}}
 \captionsetup{font={small,stretch=1.25}, labelfont={bf}}

 \renewcommand{\arraystretch}{1.2}
   \resizebox{0.475\textwidth}{!}{
  \begin{tabular}{l||c c c c c}
   \toprule[1.5pt]
    \multirow{2}{*}{\textbf{Methods}}  & \multicolumn{5}{c}{\texttt{Office-Home Dataset}} \\
    \cline{2-5}\cline{6-6}
    & \textbf{ $\cup \rightarrow$ A} & \textbf{$\cup  \rightarrow$  p} & \textbf{$\cup  \rightarrow$  c} & \textbf{$\cup  \rightarrow$ R}  & \textbf{Mean}\\
   \hline
   \hline
   -UDN& 82.82&	89.77&	72.97&	89.88& 83.86\\
   -IUS& 81.71&	87.41&	84.19&	91.81&86.28\\
   -CDC& 85.12&	90.63&	85.34&	94.10&88.80\\
   \toprule[1pt]
   \rowcolor{gray!40} \textbf{\method{} (Ours)} & \textbf{86.03}& \textbf{91.78}& \textbf{86.91}& \textbf{95.22}& \textbf{89.99}\\
   \toprule[1.5pt]
  \end{tabular}}
  \label{tab:aba}
\end{center}
\vspace{-0.4cm}
\end{table}

\begin{table}
    \begin{center}
    \caption{\textbf{LPS vs. GPG on \texttt{Office-Home} dataset.}}
 \captionsetup{font={small,stretch=1.25}, labelfont={bf}}
 \renewcommand{\arraystretch}{1.2}
   \resizebox{0.43\textwidth}{!}{
  \begin{tabular}{c||c c c c c}
   \toprule[1.5pt]
    \multirow{2}{*}{\textbf{Methods}}  & \multicolumn{5}{c}{\texttt{Office-Home Dataset}} \\
    \cline{2-5}\cline{6-6}
    & \textbf{ $\cup \rightarrow$ A} & \textbf{$\cup  \rightarrow$  p} & \textbf{$\cup  \rightarrow$  c} & \textbf{$\cup  \rightarrow$ R}  & \textbf{Mean}\\
   \hline
   \hline
   LPS& 85.60&	90.13	&85.91&	93.78&88.86
\\
   GPG & \textbf{86.03}& \textbf{91.78}& \textbf{86.91}& \textbf{95.22}& \textbf{89.99}\\
   \toprule[1.5pt]
  \end{tabular}}
  \label{tab:vs}
    \end{center}
    \vspace{-0.4cm}

\end{table}


\subsection{Dataset and Setting}
\label{sec:setting}
\noindent\textbf{Benchmark Datasets.} 
\texttt{Office-Home}~\cite{venkateswara2017deep} benchmark contains 65 classes, with 12 adaptation scenarios constructed from 4 domains (\emph{i.e.}, \textbf{R}: Real world, \textbf{C}: Clipart, \textbf{A}: Art, \textbf{P}: Product). 
\texttt{miniDomainNet} is a subset of DomainNet~\cite{peng2019moment} and contains 140,006
96×96 images of 126 classes from four domains: Clipart, Painting, Real, and Sketch (abbr. \textbf{R}, \textbf{C}, \textbf{P} and \textbf{S}). 
\texttt{Digits-five} contains five-digit sub-datasets: 
MNIST (mt)~\cite{lecun1998gradient} , Synthetic (sy)~\cite{ganin2015unsupervised}, MNIST-M(mm)~\cite{ganin2015unsupervised},
SVHN (sv)~\cite{netzer2011reading}, and USPS (up)~\cite{ganin2015unsupervised}. Each sub-dataset contains images of numbers ranging from 0 to 9.


\noindent\textbf{Implementation Details.} We employ the ResNet~\cite{he2016deep} as the backbone model on three datasets. We train \method{} with the SGD~\cite{bottou2010large} optimizer in all experiments.
Besides, we use an identical set of hyperparameters ($B$=64, $M_o$=0.9, $W_d$=0.00005, $L_r$=0.0004)\footnote{$B$, $M_o$, $W_d$ and $L_r$ refer to batch size, momentum, weight decay and learning rate in SGD optimizer.} across all datasets. Following ~\cite{xie2022active}, the total labeling budget $\mathcal{B}$ is set as 5\% of target samples, which is divided into 5 selection steps, \emph{i.e.}, the labeling budget in each round is b = $\mathcal{B}$/5. 

\begin{figure}[t]
\includegraphics[width=0.475\textwidth]{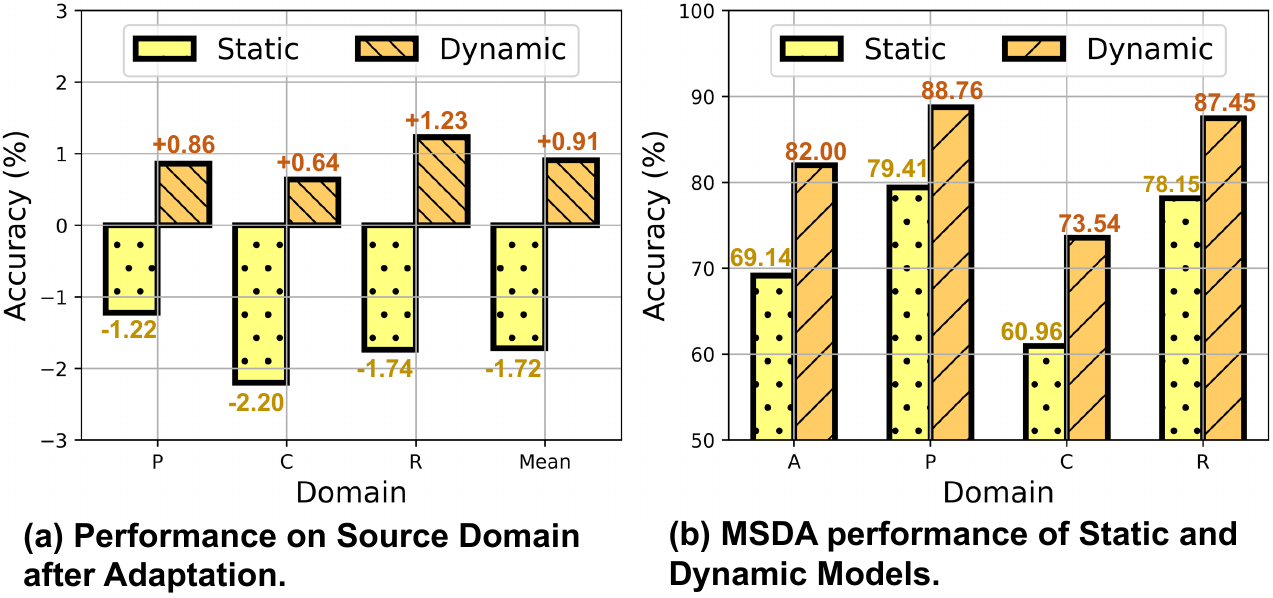}
\vspace{-0.5cm}
\centering\caption{\textbf{Performance of static and dynamic models.}}
\vspace{-0.4cm}
\label{fig:s_vs_d}
\end{figure}
\noindent\textbf{Comparison of Methods.} For quantifying the efficacy of the proposed framework, we compare \method{} with previous SoTA MSDA and ADA approaches. For MSDA methods, wo choose MCD~\cite{saito2018maximum}, DCTN~\cite{xu2018deep}, M$^3$SDA~\cite{peng2019moment}, MME~\cite{saito2019semi}, DEAL~\cite{zhou2021domain}, MFSAN~\cite{zhu2019aligning}, MDDA~\cite{zhao2020multi}, SImpAI~\cite{venkat2020your}, MADAN~\cite{zhao2021madan} as baselines. The ADA methods include CLUE~\cite{prabhu2021active}, TQS~\cite{fu2021transferable}, DUC~\cite{xie2023dirichlet} and EADA~\cite{xie2022active}\footnote{For ADA methods, we only compare with the approaches that have open source code and run them on our MADA setting.}.

\subsection{Overall Performance}
Table~\ref{tab:results_1} summarizes the quantitative MADA results of our framework and baselines on \texttt{Office-Home} (The experimental results of  \texttt{miniDomainNet} and \texttt{Digits-five} in the Appendix). We make the following observations: 
1) In general, irrespective of the different shot scenarios, compared to SoTAs, \method{} achieves the best performance on almost all the adaptation scenarios. 
In particular, \method{} outperforms other baselines in terms of mean accuracy by a large margin (\texttt{Office-Home}: \textbf{\underline{6.96\% $\sim$ 19.59\%}}) for the MADA task. 
2) It is worth noting that almost all ADA models generally outperform MSDA baselines by employing an appropriate annotation strategy, showing the necessity of considering informative target supervision. The two tables also suggest that the clustering-based ADA methods (\emph{e.g.}, CLUE) seem to be less effective than uncertainty-based methods (\emph{e.g.}, TQS, DUC, EADA) on the large-scale dataset. This may be because clustering becomes more difficult with multi-source domain modeling.
3) Benefiting from the carefully designed Integrated Uncertainty Selector (INS), our \method{} achieves better estimation of uncertainty by incorporating the distribution interpretation into prediction, while other uncertainty-aware methods can easily be miscalibrated. Further, combined with dynamic multi-domain modeling and contextual diversity enhancement, we obtain the best MADA performance. Overall, these results strongly demonstrate the effectiveness of our proposed \method{} under the practical multi-source active
domain transfer setting.

\subsection{Ablation Study}
\label{sec:aba}
\noindent\textbf{Effectiveness of Each Component.} 
We conduct an ablation study, as illustrated in Table~\ref{tab:aba}, to demonstrate the effectiveness of each component. Comparing \method{} and \method{}(-UDN) (Row 1 \emph{vs.} Row 4), the UDN contributes a 6.13\% improvement in mean accuracy. The results in Row 2 show that the IUS leads to a 3.71\% increase in mean accuracy. Meanwhile, Row 3 indicates a noticeable performance degradation of 1.19\% without the CDC. In summary, each module's contribution to improvement is distinct. When combining all components, our \method{} framework exhibits steady improvement over the baselines.



\noindent \textbf{Integration Coefficient $\lambda_{dom}$ and $\lambda_{pre}$.} 
\label{sec:hyp}
We investigate the impact of integration coefficient $\lambda_{dom}$and $\lambda_{pre}$, which controls the uncertainty integration ratio to select valuable samples. 
\begin{figure}[t]
\includegraphics[width=0.475\textwidth]{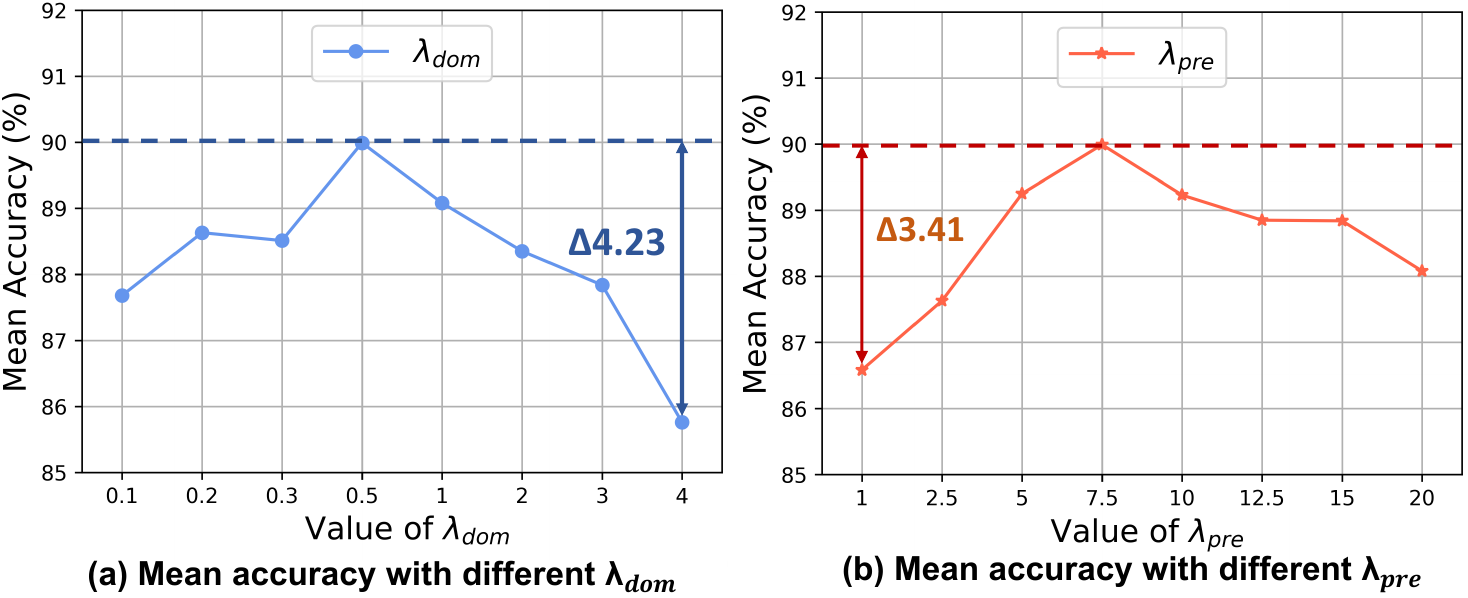}
\vspace{-0.3cm}
\centering\caption{\textbf{Ablation with respect to
$\lambda_{dom}$ and $\lambda_{pre}$.}}
\label{fig:hy}
\end{figure}
The mean accuracy of different coefficients of $\lambda_{dom}$ and $\lambda_{pre}$ on \texttt{Office-Home} dataset is shown in Figure ~\ref{fig:hy}. This figure suggests that the optimal choices of $\lambda_{dom}$ and $\lambda_{pre}$ are approximately $7.5$ and $0.5$, respectively. This phenomenon implies that domain uncertainty predominates in uncertainty estimation, which demonstrates the importance of Evidential Deep Learning (EDL) for sample selection in the presence of domain discrepancies.

\noindent \textbf{LPS \emph{vs.} GPG.} We evaluate two dynamic layer learning approaches: local parameters shift (LPS) and global parameters generation (GPG), as shown in Table~\ref{tab:vs}. The results in the table indicate that utilizing GPG yields superior MADA outcomes. Our intuition behind this observation is that GPG operates the dynamic parameters within a larger semantic space compared to LPS, thereby benefiting the modeling of complex multiple domains.

\begin{figure}[t]
\includegraphics[width=0.475\textwidth]{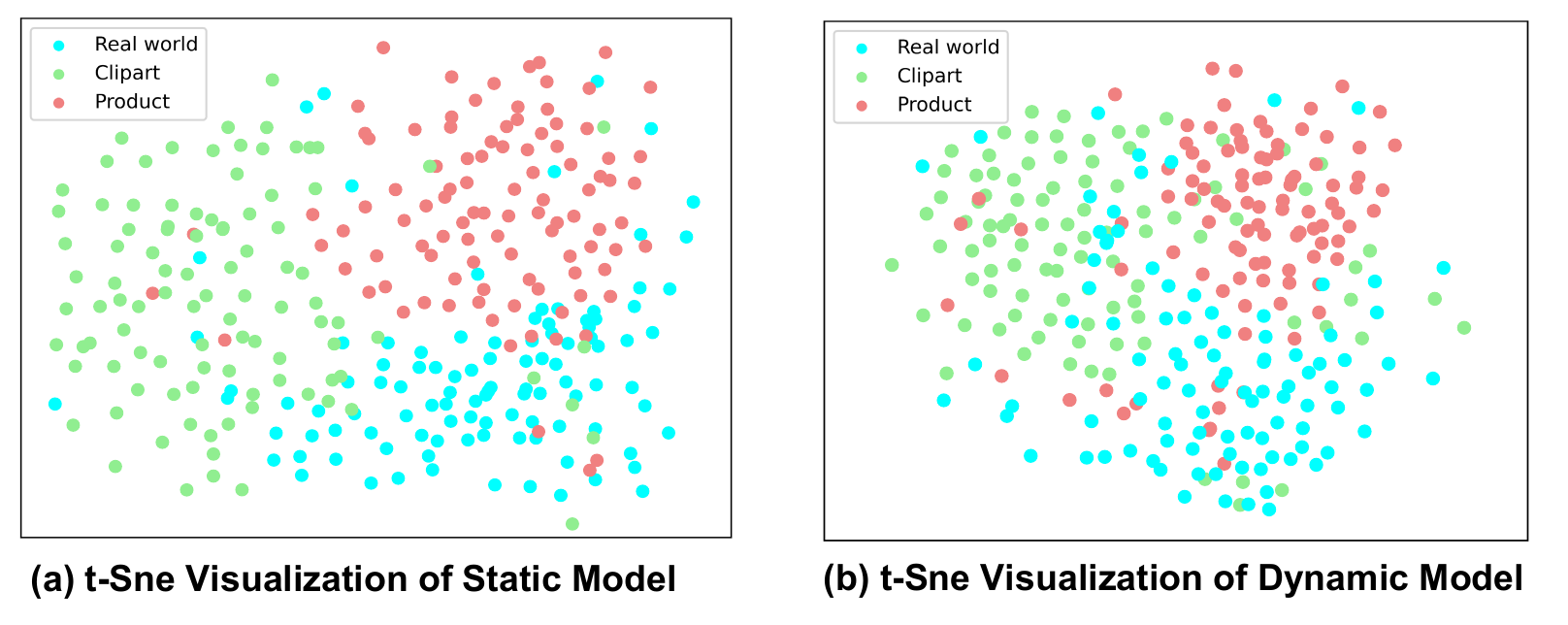}
\vspace{-0.5cm}
\centering\caption{\textbf{t-SNE Visualization of multi-source embeddings.}}
\vspace{-0.3cm}
\label{fig:st-tsne}
\end{figure}

\subsection{In-Depth Analysis}
We further validate several vital issues of the three modules, in proposed \method{} by answering the following questions.

\begin{sloppypar}
\noindent\textbf{Q1: Can the UDN appropriately model the multi-domain by a single model?} To build the insight on the effectiveness of the dynamic multi-domain learning in \method{}, for the $\cup \rightarrow \textbf{A}$ on \texttt{Office-Home} dataset, we first evaluate the performance of multi-source domain after adaptation in Figure~\ref{fig:s_vs_d}(a). A clear performance degradation (1.72\% in average) exists when using the static model, indicating the static transfer essentially averages the domain conflicts and thus the performance drops on each source domain. In contrast, our proposed dynamic model handles the domain shifts well, \emph{i.e.}, adapting to the target domain further improves the accuracy on the source domains. Figure~\ref{fig:s_vs_d}(a) reports MSDA results without target supervision, which shows that our dynamic transfer obtains a superior adaptation result against the static model, even comparable compared with ADA models. We also visualize the samples with one specific class (``chair'') on the other source domains. This figure suggests the UDN tends to group the samples with the same class in one cluster, which validates our claim that adapting the model across domains can be achieved by adapting the dynamic model to samples. However, the sample density using the static model is relatively low, since forcing a static model to leverage multi-domains may degrade the performance.
\end{sloppypar}

 \begin{figure}[t]
\includegraphics[width=0.475\textwidth]{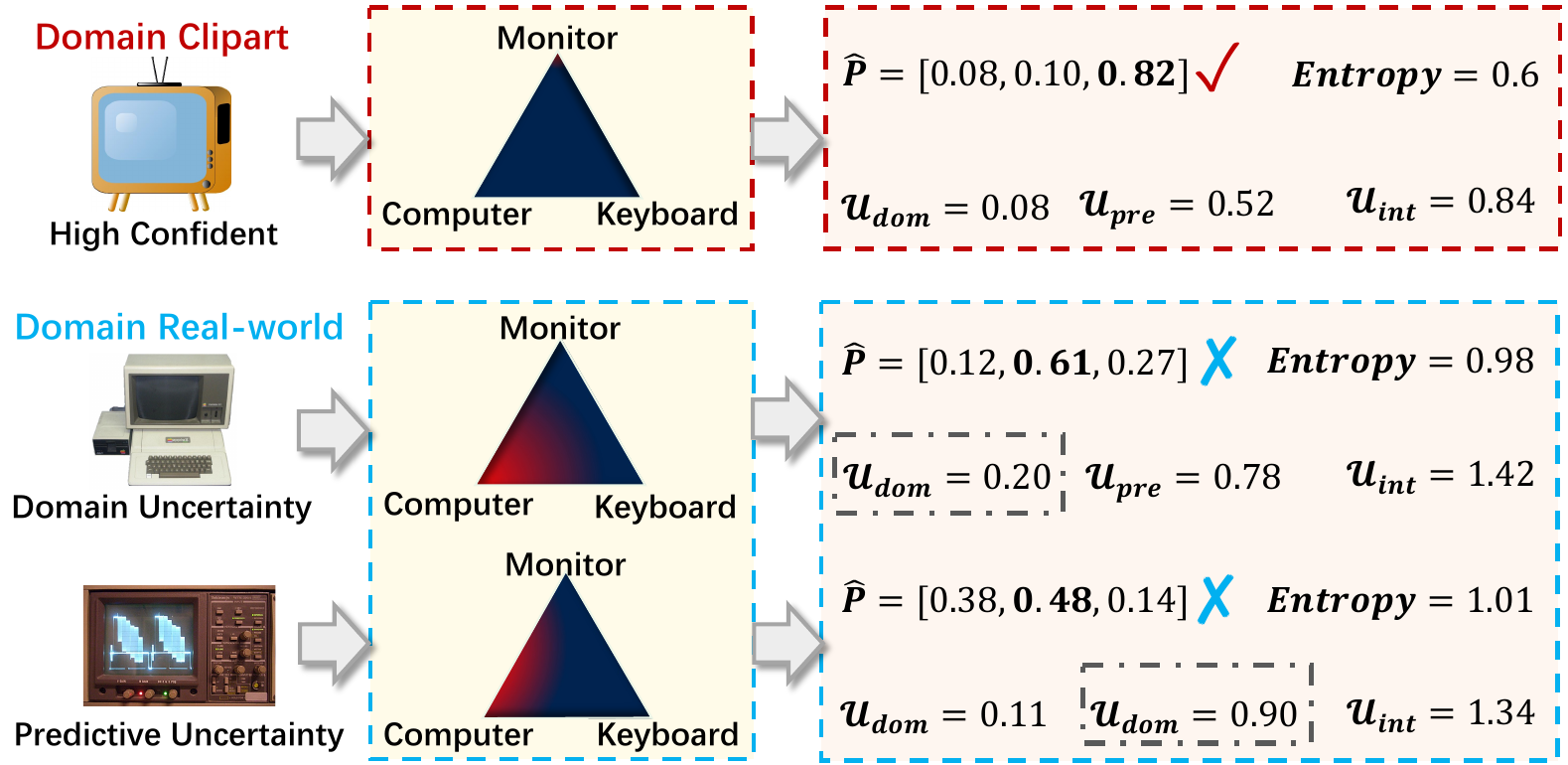}
\centering\caption{\textbf{Case study with respect to $\mathcal{U}_{dom}$ and $\mathcal{U}_{pre}$.}}
\label{fig:case}
\vspace{-0.3cm}
\end{figure}
\begin{sloppypar}
\noindent\textbf{Q2: How does the IUS calibrate the uncertainty to select informative samples?} The key insight of uncertainty calibration is to use the EDL to measure the ``targetness'', \emph{i.e}, domain uncertainty. We plot the distribution of $\mathcal{U}_{dom}$ in Figure.1 (in the Appendix), where the model is trained on the multi-source domain. We see that the $\mathcal{U}_{dom}$ of target data is noticeably biased from multi-source domain. Such results show that our $\mathcal{U}_{dom}$ can play the role of domain discriminator without introducing it. 
We also give an intuitive case study in Figure~\ref{fig:case}: given a ``monitor'', for the two images from the Real-World domain, the general prediction entropy cannot distinguish them, whereas $\mathcal{U}_{dom}$ and $\mathcal{U}_{pre}$ calculated based on the prediction distribution can reflect what contributes more to their uncertainty and be utilized to guarantee the informativeness of selected data. Additionally, we study the effect of $\mathcal{U}_{dom}$ and $\mathcal{U}_{pre}$ in IUS, we train the model with the individual uncertainty under MADA. As shown in Table.~\ref{tab:ius}, both the $\mathcal{U}_{dom}$ and $\mathcal{U}_{pre}$ play significant roles in IUS. Notably, simply using $\mathcal{U}_{pre}$ results in a higher performance decay compared to only using $\mathcal{U}_{dom}$, which implies a standard DNN can easily produce overconfident but wrong predictions for target data, making the estimated predictive uncertainty unreliable. On the contrary, reasonably integrating the $\mathcal{U}_{dom}$ and $\mathcal{U}_{pre}$ can calibrate uncertainty to choose the target samples, thereby facilitating MADA.
\end{sloppypar}
\begin{table}[t]
  \centering
    \caption{\textbf{Two uncertainties in IUS.}}
 \renewcommand{\arraystretch}{1.2}
 \resizebox{0.475\textwidth}{!}{
{
    \begin{tabular}{cc|cccc|c}
    \toprule[1pt]
    \multicolumn{2}{c|}{{\textbf{Uncertainty}}} & \multicolumn{5}{c}{\texttt{Office-Home Dataset}}\\
    \cline{1-2}\cline{3-7}
    $\mathcal{U}_{dom}$ & $\mathcal{U}_{pre}$ & \textbf{ $\cup \rightarrow$ A} & \textbf{$\cup  \rightarrow$  P} & \textbf{$\cup  \rightarrow$  C} & \textbf{$\cup  \rightarrow$ R}& \textbf{Mean} \\
    \midrule[1pt]
    \midrule[1pt]
      &\faCheckCircle  & 83.81& 88.36& 84.43& 91.13& 86.93\\
         \faCheckCircle& & 85.39 &90.63 &86.05& 93.77& 88.96 \\
    \midrule[1pt]
    \rowcolor{gray!40} \faCheckCircle  &\faCheckCircle & \textbf{86.03}& \textbf{91.78}& \textbf{86.91}& \textbf{95.22}&\textbf{89.99}\\
    \bottomrule[1pt]
    \end{tabular}
    }
}
\label{tab:ius}
\end{table}


\begin{sloppypar}
\noindent$\textbf{Q3: How does the CDC further benefit the MADA?}$ Section~\ref{sec:aba} of the ablation study has initially shown that the CDC can boost the MADA performance on \texttt{Office-Home} dataset. 
\begin{figure}[t]
    \includegraphics[width=0.475\textwidth]{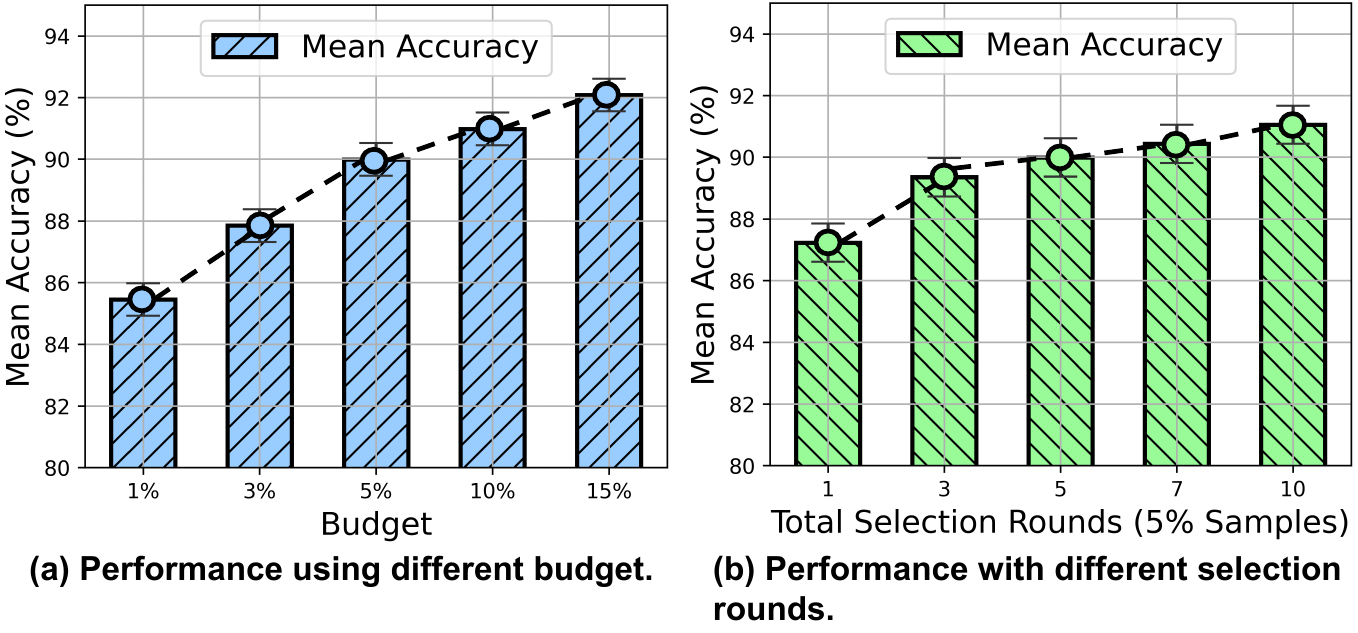}
    \vspace{-0.2cm}
\centering\caption{\textbf{Ablation study of different AL setting.}}
\label{fig:al}
\vspace{-0.3cm}
\end{figure}
To further analyze the source of the CDC's improvements, we visualize our method's sample selection behavior using t-SNE~\cite{van2008visualizing} and report the performance after the first round in Figure.2 (in the Appendix). This visualization result suggests that our \method{} tends to select representative uncertain samples with lower image-level density compared with \method{} (w/o CDC). These samples can propagate representative label information to nearby samples, avoiding the curse of scale by leveraging uncertainty to obtain preliminary samples, thereby boosting MADA performance. We also perform the hyperparameter analysis of $\lambda_u$ and $t^{\tau}$ in the Appendix.
 The observations and analysis verify the effectiveness of the CDC in improving the diversity of sample selection and thus boosting MADA performance.
\end{sloppypar}



\section{Conclusion}
\label{sec:conclusion}

We propose a novel task, termed Multi-source Active Domain Adaptation (MADA), and have correspondingly developed \textbf{Detective}. This framework comprehensively considers domain and predictive uncertainty, and context diversity to select informative target samples, thereby boosting multi-source domain adaptation. We hope that \method{} can provide new insights into the domain adaptation community.

{
    \small
    \bibliographystyle{ieeenat_fullname}
    \bibliography{reference}
}


\end{document}